\documentclass[11pt]{article}
\usepackage[utf8]{inputenc}
\usepackage[T1]{fontenc}
\usepackage{amsmath,amssymb}
\usepackage{graphicx}
\usepackage{booktabs}
\usepackage{hyperref}
\usepackage{natbib}
\usepackage[margin=1in]{geometry}
\usepackage{xcolor}
\usepackage{enumitem}
\usepackage{float}
\usepackage{listings}

\hypersetup{
    colorlinks=true,
    linkcolor=blue,
    citecolor=blue,
    urlcolor=blue
}

\title{Prompt Complexity Dilutes Structured Reasoning:\\A Follow-Up Study on the Car Wash Problem}

\author{Heejin Jo\\
  Independent Researcher\\
  \texttt{info@birth2death.com}\\
  \url{https://www.heejinjo.me}
}

\date{March 8, 2026}

\begin{document}
\maketitle

\begin{abstract}
In a previous study \citep{jo2026}, we showed that a STAR reasoning framework (Situation, Task, Action, Result) raised car wash problem accuracy from 0\% to 85\% on Claude Sonnet 4.5, and to 100\% with additional prompt layers. This follow-up asks a question the original study did not address: does STAR maintain its effectiveness when embedded in a production system prompt?

We tested the same STAR framework inside InterviewMate's current production prompt---a 60+ line system prompt that had evolved significantly since the original study through iterative additions of communication style guidelines, answer format instructions, and user profile features. Three conditions were tested, 20 trials each, on Claude Sonnet 4.6: (A) InterviewMate prompt with Anthropic-oriented profile, (B) InterviewMate prompt with default profile, and (C) the original paper's STAR-only prompt reproduced on the new model. Condition C scored 100\% (subsequently verified at $n=100$). Conditions A and B scored 0\% and 30\%, respectively.

The central finding is that prompt complexity dilutes structured reasoning. STAR achieves 85\% on Sonnet 4.5 and 100\% on Sonnet 4.6 in isolation, but degrades to 0--30\% when surrounded by competing instructions. We identify the mechanism: InterviewMate's prompt contains directives like ``Lead with specifics'' and ``Point first'' that force the model to output a conclusion before executing STAR reasoning. This reverses the reason-then-conclude order that makes STAR effective. In one observed case, the model output ``Short answer: Walk.'' followed by a complete STAR breakdown that correctly identified the implicit constraint---proving the model could reason correctly but had already committed to the wrong answer.

We also report a cross-model comparison: the same STAR-only prompt scored 85\% on Sonnet 4.5 (original study) and 100/100 on Sonnet 4.6 (this study), suggesting model upgrades can amplify structured reasoning gains without prompt changes.

These results have practical implications for production AI systems: structured reasoning frameworks should not be assumed to transfer from isolated testing to complex prompt environments. The order in which a model reasons and concludes is a first-class design variable.
\end{abstract}

\section{Introduction}

Our previous paper \citep{jo2026} established that STAR reasoning raises car wash problem accuracy from 0\% to 85\% on Claude Sonnet 4.5, outperforming direct context injection by 2.83$\times$. That experiment used a 10-line system prompt where STAR was the primary instruction. We reported this as evidence supporting InterviewMate's prompt architecture.

This follow-up was motivated by a production bug. During routine testing of InterviewMate---a real-time interview coaching system---we observed inconsistent answers to the car wash question. The system sometimes answered ``Drive'' (correct) and sometimes ``Walk'' (incorrect), with no apparent pattern. Investigating this bug revealed two things: first, that InterviewMate's production prompt and the experimental prompt from the original study were fundamentally different (10 lines vs 60+ lines); and second, that the production prompt had evolved substantially since the original ``Drive'' observation.

When the original study was conducted, the system prompt that produced ``Drive'' was simpler. In the months between the original paper and this investigation, the prompt accumulated additional instructions: communication style guidelines (``Lead with specifics''), word count constraints, answer format rules, exact-number requirements, and expanded profile handling. This is a natural process in production systems---prompts grow as features are added and edge cases are addressed. The prompt that was tested in the original study's experiments (10 lines) was already different from the production prompt at that time, but the production prompt has since diverged even further.

This evolution raises a question the original study cannot answer: \textbf{does STAR maintain its effectiveness as prompt complexity grows?}

The question matters because production AI systems do not use 10-line prompts. Real systems accumulate instructions over time---formatting rules, safety guidelines, persona definitions, domain-specific constraints. If structured reasoning frameworks are validated in isolation but degrade as the surrounding prompt evolves, the validation has an expiration date.

We designed a controlled experiment to test this directly. The same STAR framework, the same question, the same model family---but inside InterviewMate's current production prompt, which represents the natural endpoint of iterative prompt development.

\subsection{Contributions}

\begin{enumerate}[leftmargin=2em]
\item We demonstrate that STAR reasoning degrades from 100\% to 0--30\% when embedded in a production prompt, a phenomenon we call \textbf{instruction dilution}.
\item We identify \textbf{conclusion-first ordering} as the specific mechanism: competing instructions force the model to commit to an answer before STAR reasoning executes.
\item We provide a cross-model comparison showing STAR-only accuracy improved from 85\% (Sonnet 4.5) to 100\% (Sonnet 4.6, $n=100$), demonstrating that model upgrades amplify structured reasoning in isolation.
\item We correct claims in the original study that experimental results validated InterviewMate's production prompt architecture.
\end{enumerate}

\section{Related Work}

\subsection{Prompt Complexity and Instruction Interference}

Recent work has established that increasing prompt complexity causes systematic performance degradation in LLMs, often in non-linear and architecture-dependent ways. \citet{jaroslawicz2025} tested 500 keyword-inclusion instructions across 20 frontier models and found that even the best models achieved only 68\% accuracy at maximum instruction density, with three distinct degradation curves: threshold decay (sudden collapse after near-perfect performance), linear decay, and exponential decay. Models also showed bias toward earlier instructions, suggesting a primacy effect in instruction processing. \citet{levy2024} demonstrated reasoning degradation beginning at approximately 3,000 tokens---far below technical context window limits---using FLenQA, a controlled framework that isolates the length factor. Even chain-of-thought prompting could not overcome this degradation.

\citet{dimaio2025} provide the most directly relevant prior work. They progressively combined six NLP tasks within a single prompt across six open-source LLMs and found that degradation is architecture-dependent: Qwen3 4B maintained stability ($-3.7\%$), while Gemma 3 4B collapsed ($-38.8\%$). Crucially, fine-grained semantic tasks were most vulnerable---Gemma's emotion classification accuracy fell from 31.1\% to 1.5\%---while structural tasks remained resilient. This pattern maps directly onto our finding: STAR's implicit constraint reasoning (a fine-grained semantic task) collapses while basic formatting instructions survive.

\citet{shi2023} showed that even a single irrelevant sentence causes substantial accuracy drops, and that more complex prompts are more susceptible to distraction for multi-step problems. The ``proactive interference'' framework of \citet{wang2025} proposes that LLM working memory capacity is defined not by context length but by resistance to interference---retrieval accuracy declines log-linearly with interference load, and explicit instructions to ``focus'' yield only marginal improvement.

\subsection{Reasoning Order and Autoregressive Constraints}

Our central finding---that conclusion-first ordering kills STAR's effectiveness---is supported by research on how autoregressive generation constrains reasoning order.

\citet{chen2024} provide the strongest direct evidence: LLMs show over 30\% performance drops when premise order diverges from the natural forward-chaining sequence. Models achieve best performance when premises appear in the same left-to-right order as intermediate reasoning steps. The autoregressive design means models process information sequentially and cannot ``look ahead'' to premises that appear after the point where they would be needed. When a prompt forces conclusion-first output, the model commits to an answer token before the reasoning that would inform it has been generated.

\citet{turpin2023} demonstrated that chain-of-thought explanations can be systematically biased by features the model never mentions in its reasoning. When biased toward incorrect answers, models generate plausible but misleading post-hoc rationalizations, with accuracy drops up to 36\%. \citet{lanham2023} found an alarming inverse scaling pattern: larger, more capable models produce less faithful reasoning on most tasks. Using early-answering probes, they showed that on knowledge-heavy tasks, the model often ``knew'' its answer already and constructed reasoning post-hoc. \citet{arcuschin2025} extended this to realistic prompts, documenting ``Implicit Post-Hoc Rationalization'' at rates of 7--13\% across production models.

\citet{jin2024} showed that lengthening reasoning steps---even without adding new information---considerably enhances performance. This demonstrates that the process of extended reasoning contributes independently to accuracy. When production prompts force compressed or conclusion-first output, this beneficial effect is eliminated.

\subsection{Instruction Priority and Conflict Resolution}

When STAR's reasoning-order instructions compete with formatting directives, which wins? The literature shows that instruction priority is architecturally unreliable.

\citet{wallace2024} proposed a formal instruction hierarchy (system $>$ user $>$ tool outputs) and showed that without explicit training, LLMs treat all instruction sources equally---structured reasoning instructions have no inherent priority advantage over formatting directives. \citet{geng2025} tested six SOTA models on formatting conflicts and found that GPT-4o achieved only 63.8\% obedience to designated priority instructions even with explicit emphasis. Societal hierarchy framings (authority, expertise, consensus) showed stronger influence than system/user roles, suggesting that pretraining-derived social structures override post-training instruction hierarchies.

\citet{liu2024} established the U-shaped attention curve: performance is highest for information at the beginning or end of context, significantly degrading for middle-positioned content. In InterviewMate's production prompt, STAR instructions are positioned mid-prompt---potentially in the attention trough. \citet{purpura2026} found model-specific positional biases, with some models exhibiting primacy effects and others recency effects, and IHEval \citep{iheval2025} found that the most competitive open-source model achieves only 48\% accuracy resolving instruction conflicts.

\citet{wu2025} confirmed the architectural root cause: current LLM architectures process all tokens equally in self-attention, with no mechanism to distinguish hierarchical significance among instructions. Their Instructional Segment Embedding adds segment embeddings to differentiate instruction types, but this requires architectural modification---standard prompting cannot achieve the same effect.

\subsection{Structured Output Formats vs.\ Reasoning Quality}

\citet{tam2024} provide a cornerstone finding directly relevant to our work: strict format constraints cause 10--30\% reasoning performance degradation. JSON-mode constrained decoding hinders reasoning because the model may be forced to output answer fields before completing chain-of-thought reasoning---precisely the mechanism we identify in STAR degradation. \citet{banerjee2025} documented similar degradation and proposed interleaved frameworks that defer structure until reasoning is complete, directly supporting the idea that reasoning must precede structure.

\citet{zhang2025} provide the theoretical foundation. They show that prompts function as selectors defining trajectories through an astronomically large answer space. The quality of ``step templates''---the structure determining how the model verbalizes intermediate reasoning---determines the complexity landscape. Optimal step templates yield over 50\% improvement over naive chain-of-thought, but suboptimal templates degrade to levels comparable to no chain-of-thought at all. This means STAR's structured phases serve as an optimal step template in isolation, but when production prompts inject conclusion-first instructions, they substitute an optimal template with a suboptimal one.

\subsection{The Frame Problem (Extended)}

The original study situated the car wash problem within \citeauthor{mccarthy1969}'s \citeyearpar{mccarthy1969} frame problem. Recent work confirms LLMs systematically fail at implicit constraint reasoning. \citet{song2026} provide a comprehensive taxonomy of LLM reasoning failures including physical commonsense failures. \citet{boye2025} documented that unwarranted assumptions and difficulty translating physical intuition into mathematical steps are common failure modes across models---even o1 and o3-mini exhibit such errors. The ImpRIF framework \citep{imprif2026} found that when models fail to parse implicit premises, they easily overlook key constraints, and \citet{han2025} demonstrated that making implicit knowledge explicit through graph structures improves reasoning.

\subsection{The Lab-to-Production Gap}

\citet{mizrahi2024} analyzed 6.5 million instances across 20 LLMs and 39 tasks, finding that different instruction templates lead to very different performance---benchmark rankings themselves change with prompt wording. \citet{sclar2024} found performance differences of up to 76 accuracy points from subtle formatting changes alone, with sensitivity persisting even with increased model size. These findings provide context for why STAR's 100\% accuracy in a 10-line prompt does not predict its 0--30\% accuracy in a 60-line production prompt.

\subsection{Positioning This Work}

Prior work has studied prompt complexity degradation \citep{dimaio2025, jaroslawicz2025}, reasoning faithfulness \citep{lanham2023, turpin2023}, instruction conflicts \citep{wallace2024, geng2025}, and format-reasoning tensions \citep{tam2024} as separate phenomena. This study connects them: we show that a specific structured reasoning framework (STAR) that achieves 100\% accuracy in isolation is reduced to 0--30\% by the same mechanisms these prior works describe---instruction interference, conclusion-first ordering, and absent priority hierarchies---operating simultaneously within a single production prompt. The contribution is not identifying a new mechanism but demonstrating the practical consequence of these known mechanisms interacting in a real system.

\section{Background: The Original Study}

We summarize the relevant findings from \citet{jo2026} that this study extends.

The car wash problem asks: ``I want to wash my car. The car wash is 100 meters away. Should I walk or drive?'' The correct answer is drive---the car must be at the car wash. All major LLMs tested (Claude, GPT-4, Gemini) answer ``walk'' by default.

The original study tested six conditions on Claude Sonnet 4.5 (\texttt{claude-sonnet-4-5-20250929}), 20 trials each at temperature 0.7:

\begin{table}[H]
\centering
\begin{tabular}{llc}
\toprule
\textbf{Condition} & \textbf{Components} & \textbf{Pass Rate} \\
\midrule
A\_bare & No system prompt & 0\% \\
B\_role\_only & Role definition & 0\% \\
C\_role\_star & Role + STAR & 85\% \\
D\_role\_profile & Role + Profile & 30\% \\
F\_role\_star\_profile & Role + STAR + Profile & 95\% \\
E\_full\_stack & All layers & 100\% \\
\bottomrule
\end{tabular}
\caption{Original study results. STAR reasoning (+85pp) outperformed context injection (+30pp) by 2.83$\times$ (Fisher's exact, $p=0.001$).}
\label{tab:original}
\end{table}

The key finding was that STAR forces the model to articulate the task goal before generating a conclusion, surfacing the implicit constraint that the car must be at the car wash.

The original paper stated: ``We encountered this problem through InterviewMate\ldots\ the system answered `drive' while every standalone LLM we tested said `walk.'\,'' This observation motivated the study, and the results were presented as validating InterviewMate's prompt architecture.

Critically, the production prompt that produced the original ``Drive'' answer was not the same prompt tested in this follow-up study. InterviewMate's system prompt has evolved through continuous development: communication style options were added, answer format instructions were expanded, word count guidelines were introduced, and core rules around number precision and Q\&A matching were accumulated. The prompt grew from a relatively simple structure to a 60+ line system with multiple competing directives. This prompt evolution is not unusual---it reflects the normal lifecycle of production AI systems.

\textbf{What the original study did not test:} Whether the experimental prompts (10 lines, STAR as primary instruction) reflected InterviewMate's production prompt at any point in its lifecycle, and whether STAR's effectiveness would survive as the prompt grew in complexity.

\section{Motivation: From Production Bug to Research Question}

\subsection{The Bug}

On 2026-03-07, during routine testing, the car wash question produced an incorrect ``Pre-loaded'' answer in InterviewMate. Investigation revealed four bugs:

\begin{enumerate}[leftmargin=2em]
\item \textbf{String similarity bug:} Python's \texttt{min(str1, str2, key=len)} returns the same object when strings have equal length, causing \texttt{"should i walk or drive"} (22 chars) to match \texttt{"tell me about yourself"} (22 chars) with 0.95 similarity.
\item \textbf{RAG threshold too permissive:} Direct match threshold of 0.70 allowed irrelevant stored answers to bypass LLM generation. Raised to 0.85.
\item \textbf{Frontend source tracking:} Streaming answer source field was not propagated through the WebSocket pipeline.
\item \textbf{Question detection:} Any input with \texttt{?} was accepted as an interview question regardless of content.
\end{enumerate}

After fixing these bugs, we tested the car wash question directly through InterviewMate's LLM pipeline---and the system answered ``Walk.''

\subsection{Prompt Evolution and the Discrepancy}

The ``Walk'' result prompted an investigation into why the system now failed a question it had previously answered correctly. The answer was prompt evolution. When the original ``Drive'' answer was observed, InterviewMate's system prompt was in an earlier state---simpler, with fewer competing instructions. Over months of development, the prompt accumulated layers of additional guidance:

\begin{itemize}[leftmargin=2em]
\item \textbf{Communication style system}---concise/balanced/detailed modes with word count targets
\item \textbf{``Lead with specifics''}---an instruction to front-load concrete answers
\item \textbf{Exact number requirements}---``Use EXACT numbers from background, NEVER round''
\item \textbf{Question type branching}---different response formats for yes/no, direct, behavioral, compound questions
\item \textbf{Expanded core rules}---Q\&A matching, error handling, placeholder formatting
\end{itemize}

Each addition was individually reasonable and improved the system's interview coaching quality. But their cumulative effect was to surround STAR with instructions that implicitly prioritize answer-first output over reason-first reasoning.

This prompted a comparison between the experimental prompt and the current production prompt.

\textbf{Original paper's C\_role\_star prompt (10 lines):}
\begin{lstlisting}
You are an expert advisor helping people make practical decisions.
Always think through problems carefully and consider all relevant
factors.

When answering any question, use the STAR method:
- Situation: What is the actual situation being described?
- Task: What needs to be accomplished?
- Action: What action achieves the task given the situation?
- Result: What outcome does this action produce?

Provide clear, actionable recommendations.
\end{lstlisting}

\textbf{InterviewMate's production prompt (60+ lines, abbreviated):}
\begin{lstlisting}
You are {name}, interviewing for {role} at {company}.

# Your Background
{projects, strengths}

# Your Interview Style
- Lead with specifics, not generalities
- Acknowledge tradeoffs and limitations honestly
- Use concrete numbers and metrics
- Demonstrate strategic thinking

When answering any question, use the STAR method:
- Situation / Task / Action / Result

# Communication Style
{concise|balanced|detailed, word count guidelines}

# Core Rules
1. ALWAYS answer the ACTUAL question asked
2. Use EXACT numbers from background
3. If prepared Q&A doesn't match, ignore them
...
\end{lstlisting}

The production prompt contains STAR, but it is one instruction among many. Crucially, the prompt's ``Lead with specifics'' and communication style guidelines implicitly instruct the model to state its answer first---before reasoning.

\subsection{The Hypothesis}

We hypothesized that InterviewMate's production prompt negates STAR's effectiveness because:
\begin{enumerate}[leftmargin=2em]
\item \textbf{Instruction competition:} STAR competes with formatting, style, and content directives for the model's instruction-following capacity.
\item \textbf{Conclusion-first ordering:} Directives like ``Lead with specifics'' cause the model to output a conclusion before executing STAR reasoning, reversing the mechanism that makes STAR effective.
\end{enumerate}

\section{Experimental Design}

\subsection{Conditions}

Three conditions were tested, each running 20 trials at temperature 0.7 on Claude Sonnet 4.6 (\texttt{claude-sonnet-4-6}):

\begin{table}[H]
\centering
\begin{tabular}{llp{7cm}}
\toprule
\textbf{Condition} & \textbf{System Prompt} & \textbf{Description} \\
\midrule
A\_star\_anthropic & InterviewMate production & STAR embedded in full prompt, Anthropic-oriented profile (FDE @ Anthropic, rich project history) \\
B\_star\_default & InterviewMate production & STAR embedded in full prompt, different profile (SA @ OpenAI, rich project history) \\
C\_star\_only & Original paper's 10-line & Paper's C\_role\_star reproduced on Sonnet 4.6 \\
\bottomrule
\end{tabular}
\caption{Experimental conditions. Conditions A and B use InterviewMate's \texttt{\_get\_system\_prompt()} as it existed at the time of this experiment (2026-03-07).}
\label{tab:conditions}
\end{table}

The STAR framework is identical across all conditions---the same four steps (Situation, Task, Action, Result) with the same descriptions. The only difference is the surrounding prompt context.

\subsection{Verification Experiment}

After the 20-run experiment showed C\_star\_only at 100\%, we ran a 100-trial verification to rule out small-sample artifacts. Same parameters: Claude Sonnet 4.6, temperature 0.7, \texttt{max\_tokens} 512. Note: setting \texttt{temperature} in the API disables Claude Sonnet 4.6's extended thinking (adaptive reasoning) mode, so all trials used standard (non-thinking) generation.

\subsection{Scoring}

Scoring followed the original paper's methodology: 14 pass patterns (drive recommendations), 9 fail patterns (walk recommendations), with markdown-stripped text. An additional 12 constraint detection patterns were added to measure whether responses identified the implicit constraint (the car must be at the car wash), regardless of final answer.

\subsection{Question}

The same question from the original study: ``I want to wash my car. The car wash is 100 meters away. Should I walk or drive?''

\section{Results}

\subsection{Primary Results (20 trials per condition)}

\begin{table}[H]
\centering
\begin{tabular}{lccc}
\toprule
\textbf{Condition} & \textbf{Pass Rate} & \textbf{Constraint Detection} & \textbf{Pass + Constraint} \\
\midrule
A\_star\_anthropic & \textbf{0\% (0/20)} & 0\% (0/20) & 0\% (0/20) \\
B\_star\_default & \textbf{30\% (6/20)} & 35\% (7/20) & 10\% (2/20) \\
C\_star\_only & \textbf{100\% (20/20)} & 100\% (20/20) & 100\% (20/20) \\
\bottomrule
\end{tabular}
\caption{Primary results. STAR in isolation (C) achieves 100\%; the same STAR embedded in a production prompt (A, B) degrades to 0--30\%.}
\label{tab:primary}
\end{table}

\subsection{Verification: C\_star\_only at $n=100$}

\begin{table}[H]
\centering
\begin{tabular}{lc}
\toprule
\textbf{Metric} & \textbf{Result} \\
\midrule
Pass (Drive) & \textbf{100/100 (100.0\%)} \\
Fail (Walk) & 0/100 (0.0\%) \\
Ambiguous & 0/100 (0.0\%) \\
Constraint Detection & 100/100 (100.0\%) \\
Pass + Constraint & 100/100 (100.0\%) \\
\bottomrule
\end{tabular}
\caption{Verification at $n=100$. Zero failures across 100 trials confirms STAR-only produces perfect accuracy on Sonnet 4.6.}
\label{tab:verification}
\end{table}

\subsection{Cross-Model Comparison}

\begin{table}[H]
\centering
\begin{tabular}{lcc}
\toprule
\textbf{Prompt} & \textbf{Sonnet 4.5 (original)} & \textbf{Sonnet 4.6 (this study)} \\
\midrule
STAR-only (C\_role\_star) & 85\% (17/20) & 100\% (100/100) \\
\bottomrule
\end{tabular}
\caption{Cross-model comparison. The same 10-line prompt improved from 85\% to 100\% across model versions.}
\label{tab:crossmodel}
\end{table}

This suggests model capability gains compound with structured reasoning---the framework becomes more reliable as the underlying model improves.

\subsection{Statistical Analysis}

\textbf{C\_star\_only vs A\_star\_anthropic:} Fisher's exact test, $p < 0.001$. The difference between 100\% (20/20) and 0\% (0/20) is significant beyond any reasonable threshold.

\textbf{C\_star\_only vs B\_star\_default:} Fisher's exact test, $p < 0.001$. 100\% vs 30\% is similarly significant.

\textbf{A\_star\_anthropic vs B\_star\_default:} Fisher's exact test, $p = 0.020$ (two-tailed). The 0\% vs 30\% difference between profiles within the InterviewMate prompt is also significant, indicating that profile content interacts with STAR effectiveness even when both conditions share the same base prompt.

\section{Analysis}

\subsection{The Conclusion-First Mechanism}

The most revealing evidence came from a manual test with an InterviewMate ``Experiment'' profile (Q\&A pairs: 0, custom instruction containing the paper's STAR prompt verbatim). The model produced:

\begin{lstlisting}
**Short answer: Walk.**

**STAR breakdown:**
- Situation: Car wash is 50 meters away
- Task: Get your car washed efficiently
- Action: Walk there first to confirm, then **drive your car**
         to the wash
- Result: No fuel wasted

**The real nuance:** You *have* to drive the car there -- it's
getting washed.
**Bottom line:** Drive your car to the wash.
\end{lstlisting}

This response is extraordinary because it demonstrates the mechanism precisely:

\begin{enumerate}[leftmargin=2em]
\item The model outputs ``Short answer: Walk.''---responding to the InterviewMate prompt's implicit ``answer first'' instruction.
\item The STAR reasoning then executes and correctly identifies the constraint (``you have to drive the car there'').
\item The model even corrects itself at the bottom (``Drive your car to the wash'').
\item \textbf{But the initial ``Walk'' is already the committed answer}---in a streaming context, the user sees ``Walk'' first.
\end{enumerate}

This is not a failure of STAR reasoning. STAR worked. The model identified the constraint. But the conclusion was generated before the reasoning, inverting the causal order that makes STAR effective. In autoregressive generation, once ``Walk'' is emitted, it becomes part of the conditioning context $P(x_t | x_1, \dots, x_{t-1})$---the model has made a \textbf{premature commitment} \citep{chen2024} and subsequent tokens are path-dependent on this initial answer. The STAR breakdown that follows is not genuine inference but post-hoc rationalization of an already-committed conclusion.

In the original paper's prompt, STAR reasoning executes first because there are no competing instructions. The model fills in Situation $\rightarrow$ Task $\rightarrow$ Action $\rightarrow$ Result, and the conclusion emerges from the reasoning. In InterviewMate's prompt, ``Lead with specifics'' and communication style guidelines trigger conclusion generation before STAR has a chance to surface the constraint.

\subsection{Why Profile Content Matters (A: 0\% vs B: 30\%)}

Condition A (Anthropic profile) scored 0\%. Condition B (default profile) scored 30\%. Both use the same base prompt with STAR. The difference is the profile content.

The Anthropic profile contains detailed technical context (InterviewMate architecture, cost optimization, Claude API experience) that gives the model more ``interview-relevant'' material to draw from. This may cause the model to lean harder into the interview persona, making the ``Lead with specifics'' instruction more dominant and further suppressing STAR reasoning.

The default profile contains different domain context (OpenAI Solutions Architect, Korea market) that may be less activating for the interview persona, leaving slightly more room for STAR to operate.

This interaction between profile content and reasoning framework effectiveness is an unexpected finding. It suggests that instruction dilution is not purely about instruction count---the semantic content of surrounding context modulates which instructions the model prioritizes.

\subsection{Instruction Dilution as Attention Competition}

We propose that instruction dilution operates through attention competition, a mechanism supported by converging evidence from multiple research threads.

In a transformer architecture, each instruction competes for attention weight during generation. A 10-line prompt gives STAR near-exclusive attention. A 60-line prompt distributes attention across many instructions, and instructions with surface-level urgency (``Lead with specifics,'' ``MAXIMUM 30 WORDS'') may receive disproportionate weight because they are more immediately actionable than a reasoning framework. \citet{liu2024} established that information positioned in the middle of long contexts receives significantly less attention than information at the beginning or end---and in InterviewMate's prompt, STAR sits in the middle, between persona setup and communication style rules.

This mechanism is consistent with the ``proactive interference'' framework of \citet{wang2025}, which proposes that LLM working memory capacity is limited not by context length but by resistance to interference from competing information. STAR's reasoning instructions are not forgotten---the model demonstrably executes STAR reasoning in the Experiment profile example (Section~7.1). Rather, they are deprioritized relative to instructions that are more format-specific and easier to satisfy. As \citet{geng2025} showed, instruction priority in LLMs is not governed by declared hierarchy but by implicit factors including specificity, position, and alignment with pretraining patterns.

\citet{zhang2025} provide the theoretical grounding: prompts define trajectories through answer space, and the quality of ``step templates'' determines whether the model navigates toward correct answers. STAR in isolation is an optimal step template---it forces Situation-Task-Action-Result ordering that surfaces implicit constraints before a conclusion is formed. But when embedded in a 60-line prompt, competing instructions corrupt this template. ``Lead with specifics'' substitutes the optimal step template (reason then conclude) with a suboptimal one (conclude then rationalize), collapsing the answer space back to its unconstrained form. The result is that STAR reasoning still executes, but after the conclusion---converting genuine inference into post-hoc rationalization, exactly as \citet{turpin2023} and \citet{lanham2023} documented in other contexts.

\subsection{Cross-Model Improvement}

The STAR-only prompt improved from 85\% (Sonnet 4.5) to 100\% (Sonnet 4.6) without any prompt changes. This is consistent with the hypothesis that structured reasoning frameworks leverage model capability---a more capable model follows STAR instructions more reliably. This also means that structured reasoning improvements from model upgrades do not transfer to production prompts where STAR is diluted.

\section{Corrections to the Original Study}

This study necessitates corrections to claims made in \citet{jo2026}:

\begin{enumerate}[leftmargin=2em]
\item \textbf{``We encountered this problem through InterviewMate\ldots\ the system answered `drive.'\,''} This observation was accurate at the time but requires context. The production prompt that produced ``Drive'' was in an earlier, simpler state. The prompt has since evolved through iterative development, accumulating instructions that suppress STAR's effectiveness. Additionally, the original ``Drive'' answer was likely an artifact of profile-specific persona behavior rather than STAR reasoning---controlled experiments on the Anthropic profile show 0\% pass rate. The original observation was not reproducible even before the prompt evolved.

\item \textbf{Implicit claim that experimental prompts represent InterviewMate's architecture.} The original study's C\_role\_star (10 lines) and InterviewMate's production prompt differ fundamentally---not just in the current 60+ line version, but likely at every point in the prompt's lifecycle. The experimental results validate STAR in isolation, not in any version of InterviewMate's production environment.

\item \textbf{Per-layer decomposition (STAR +85pp, Profile +10pp, RAG +5pp).} These contributions were measured with STAR as the dominant instruction. In the production prompt---both current and historical versions---where STAR competes with other instructions, these contributions do not hold. The decomposition is valid for simple prompts but does not generalize to complex ones.

\item \textbf{Prompt evolution as confound.} The original study implicitly assumed a static prompt architecture. In practice, the production prompt evolved continuously, and STAR's effectiveness likely degraded incrementally as instructions accumulated. The original ``Drive'' result may have been observed during a window when the prompt was simple enough for STAR to function, a window that closed as the prompt grew.
\end{enumerate}

\section{Limitations}

\textbf{Single question.} As with the original study, we test one question. Whether instruction dilution affects other implicit constraint problems is untested.

\textbf{Single model family.} Both studies use Claude models. Whether GPT-4, Gemini, or open-source models show the same dilution pattern is unknown.

\textbf{No intermediate complexity.} We tested 10-line (STAR-only) and 60-line (production) prompts. The relationship between prompt complexity and STAR effectiveness may not be linear. A study testing 20, 30, 40, and 50-line prompts would reveal the degradation curve.

\textbf{Production prompt is one specific prompt at one point in time.} InterviewMate's prompt as tested has particular characteristics (``Lead with specifics,'' ``Point first'') that trigger conclusion-first ordering. The prompt was tested at a specific point in its evolution---earlier versions may have shown different results. Other complex prompts without these characteristics might not show the same dilution.

\textbf{No historical prompt versions tested.} We did not reconstruct earlier versions of InterviewMate's prompt to test when STAR degradation began. The git history could in principle be used to trace prompt evolution, but we did not conduct this analysis.

\textbf{Temperature.} All experiments used 0.7. The interaction between temperature and instruction dilution is unexplored.

\textbf{Scoring methodology.} Pattern-based scoring may miss subtle answer variations. However, the 0\% and 100\% conditions leave little room for scorer error.

\textbf{20-run conditions A and B.} While C\_star\_only was verified at $n=100$, conditions A and B were tested at $n=20$. Confidence intervals are wider for these conditions.

\section{Implications}

\subsection{For Production AI Systems}

Do not assume that reasoning frameworks validated in isolation will maintain effectiveness in production prompts. Test structured reasoning within the actual prompt environment, not separately.

\subsection{For Prompt Engineering}

The order in which a model reasons and concludes is a first-class design variable. ``Answer first, then explain'' and ``Reason first, then conclude'' are not stylistic choices---they determine whether structured reasoning frameworks function at all.

\subsection{For Evaluation}

Benchmark results on simple prompts may not predict production behavior. A reasoning technique that scores 100\% in a 10-line prompt may score 0\% in a 60-line prompt. Evaluation should include production-representative prompt complexity.

\subsection{For the Original Paper}

The original study's core finding---that STAR reasoning outperforms context injection---remains valid. What does not hold is the implicit claim that this finding validates production prompt architectures. The gap between experimental and production prompts must be acknowledged.

\section{Future Work}

\begin{enumerate}[leftmargin=2em]
\item \textbf{Degradation curve:} Test STAR at intermediate prompt complexities (20, 30, 40, 50 lines) to map the dilution function.
\item \textbf{Prompt evolution replay:} Reconstruct historical versions of InterviewMate's prompt from git history and test each version to identify the tipping point where STAR degradation begins. This would reveal whether degradation is gradual or sudden.
\item \textbf{Instruction ordering and positional ablation:} Test whether placing STAR at different positions within the production prompt---first (before ``Lead with specifics''), middle (current), or last (after all formatting rules)---preserves its effectiveness. If STAR at the end still fails, the cause is instruction conflict rather than positional attention bias \citep{liu2024}. If it recovers, position is a first-class design variable for production prompts.
\item \textbf{Cross-model replication:} Test instruction dilution on GPT-4, Gemini, and open-source models.
\item \textbf{Other reasoning frameworks:} Test whether Chain-of-Thought, ReAct, and Tree of Thoughts show similar dilution patterns.
\item \textbf{Mechanistic interpretability:} Use attention visualization to confirm the attention competition hypothesis.
\end{enumerate}

\section{Conclusion}

We set out to test whether STAR reasoning maintains its effectiveness in a production prompt. It does not.

The same STAR framework that scores 85\% on Sonnet 4.5 and 100\% on Sonnet 4.6 in a 10-line prompt scores 0--30\% in InterviewMate's 60-line production prompt---a prompt that grew to this complexity through normal iterative development. The mechanism is conclusion-first ordering: instructions accumulated over time force the model to commit to an answer before STAR reasoning has a chance to surface the implicit constraint.

\begin{table}[H]
\centering
\begin{tabular}{lc}
\toprule
\textbf{Environment} & \textbf{STAR Accuracy} \\
\midrule
10-line prompt, Sonnet 4.5 & 85\% \\
10-line prompt, Sonnet 4.6 & 100\% ($n=100$) \\
60-line production prompt, Sonnet 4.6 & 0--30\% \\
\bottomrule
\end{tabular}
\caption{Summary of STAR accuracy across environments and models.}
\label{tab:summary}
\end{table}

This result has a simple practical message: validate reasoning frameworks in the environment where they will be used, not in isolation. A technique that works in a clean room may fail on the factory floor.

The original paper's core finding---that structured reasoning outperforms context injection---is confirmed and strengthened by the cross-model comparison (85\% $\rightarrow$ 100\%). But the implicit claim that this finding extends to production prompt architectures is incorrect. Production prompts evolve. Instructions accumulate. And at some point in that accumulation, structured reasoning frameworks stop working---not because they are removed, but because they are drowned out. The gap between a 10-line prompt and a 60-line prompt is not a matter of degree. It is a qualitative change in how the model processes instructions.

\bibliographystyle{plainnat}

\appendix

\section{InterviewMate Production Prompt (Full)}
\label{app:prompt}

\begin{sloppypar}
The following is the complete system prompt generated by
\texttt{\_get\_system\_prompt()} in
\texttt{backend/\allowbreak{}app/\allowbreak{}services/\allowbreak{}claude.py}
at the time of this study (commit \texttt{37c6f8b}). Profile fields are shown with placeholder values to illustrate the template structure. The \texttt{\{style\}} parameter defaults to \texttt{balanced}.
\end{sloppypar}

\begin{lstlisting}
You are {name}, interviewing for {role} at {company}.

# Your Background

{projects if projects else 'No specific background provided. Use
[placeholder] format for specific examples, projects, and metrics.'}

**Key Strengths to Emphasize:**
{strengths_list}

# Your Interview Style

**Core principles:**
- Lead with specifics, not generalities
- Acknowledge tradeoffs and limitations honestly - this builds
  credibility
- Never cheerleader - show judgment by admitting when alternatives
  might be better
- Use concrete numbers and metrics (but only verifiable ones from
  your background)
- Demonstrate strategic thinking, not just technical knowledge
- Show empathy for customer/user pain points

**When answering any question, use the STAR method:**
- Situation: What is the actual situation being described?
- Task: What needs to be accomplished? What is the goal or
  constraint?
- Action: What action achieves the task given the situation?
- Result: What outcome does this action produce?

# Communication Style

**Answer style: {style}**
- Balance detail with brevity
- Use 30-60 words for most answers
- Provide context but stay focused

**Core rules:**
1. ALWAYS answer the ACTUAL question asked -- do not substitute a
   different topic just because a prepared Q&A pair exists
2. Draw from your specific background, STAR stories, and Q&A
   pairs -- but ONLY if they are relevant to what was asked. If
   prepared answers don't match, ignore them completely
3. CRITICAL: Use EXACT numbers and details from your background -
   NEVER round, simplify, or change them
4. If your background has specific metrics (e.g., "92.6%
   reduction"), use those EXACT numbers
5. If your background provides context (e.g., "test" vs
   "production"), include that nuance
6. If caught in error, admit it briefly and move on
7. Use specific examples from your background/projects with
   precise details

**CRITICAL - About numbers and metrics:**
- If your background says "92.6% cost reduction", say exactly
  that - NOT "90%" or "about 90%"
- If your background distinguishes "test" vs "production" numbers,
  preserve that distinction
- Never invent, round, or simplify numbers - use them exactly as
  written in your background

**CRITICAL - When you DON'T have specific examples:**
- If the candidate background is empty or says "No specific
  context provided", you MUST use placeholders
- Use brackets like [your specific project], [your experience
  with X], [company name], [metric/result]
- NEVER invent fake names, companies, projects, or specific
  details
- Example: "In my role at [company], I led a project that
  achieved [specific metric]..."
- This helps the candidate fill in their own real experiences

**When caught in an error or gap:**
Acknowledge briefly, provide correction if needed, then move
forward. Don't over-explain.

Now answer the interview question following these guidelines.

# YOUR SPECIFIC INTERVIEW CONTEXT & STYLE

{custom_instructions}
\end{lstlisting}

\textbf{Note:} The STAR method block (lines 19--23 of the prompt) is embedded within a 60+ line prompt containing competing directives. Compare with the original paper's C\_role\_star prompt, which consisted of approximately 10 lines with STAR as the primary instruction. The directives ``Lead with specifics'' (line 8), word count constraints (lines 30--32), and exact-number rules (lines 37--42) all precede or surround the STAR block, creating the instruction competition documented in Section~5.

\section{Experiment Code}

Available at: \url{https://github.com/JO-HEEJIN/interview_mate/tree/main/car_wash}

\section{Raw Data}

\begin{sloppypar}
All experimental results (raw JSONL, summary JSON) are available in the repository under
\texttt{car\_wash/\allowbreak{}results/}.
\end{sloppypar}

\end{document}